%% file: iclr2027_conference.tex
\documentclass{article} 
\usepackage{iclr2027_conference,times}

\input{math_commands.tex}

\usepackage{hyperref}
\usepackage{url}
\usepackage{booktabs,multirow,graphicx,adjustbox,threeparttable}
\usepackage{float}
\usepackage{wrapfig}
\usepackage{placeins}
\usepackage{needspace}
\usepackage{amsmath,amssymb}
\usepackage{xcolor}
\usepackage{colortbl}
\usepackage{tikz}
\usepackage{tabularx}
\usepackage[skins,breakable]{tcolorbox}
\definecolor{swenextblue}{HTML}{3A6EA5}
\definecolor{swenextbluebg}{HTML}{F2F7FC}
\newtcolorbox{swenextbox}[2][]{%
  enhanced,
  breakable,
  colback=swenextbluebg,
  colframe=swenextblue!75!black,
  colbacktitle=swenextblue!18,
  coltitle=black,
  fonttitle=\bfseries,
  before upper={\raggedright},
  boxrule=0.8pt,
  arc=2mm,
  left=1.4mm,
  right=1.4mm,
  top=1.0mm,
  bottom=1.0mm,
  title={#2},
  #1
}
\definecolor{mainGPT}{HTML}{E5EDF7}
\definecolor{mainClaude}{HTML}{E5F0E6}
\definecolor{mainOurs}{HTML}{FFF1E5}
\definecolor{mainAverage}{HTML}{EFEFEF}
\usetikzlibrary{arrows.meta,positioning,shapes.misc}

\colorlet{regeOurs}{swenextblue!10}
\colorlet{regeModelName}{swenextblue!75!black}
\newcommand{\papermainstyle}{%
  \small
  \papermaincaptionstyle
  \setlength{\tabcolsep}{4pt}%
  \renewcommand{\arraystretch}{1.06}%
  \setlength{\aboverulesep}{0.4ex}%
  \setlength{\belowrulesep}{0.65ex}%
  \setlength{\abovecaptionskip}{0pt}%
  \setlength{\belowcaptionskip}{5pt}}
\newcommand{\papermainnotes}{%
  \footnotesize
  \setlength{\itemsep}{2pt}%
  \renewcommand{\makelabel}[1]{}}
\newcommand{\papermodelname}[1]{\textcolor{regeModelName}{\textbf{#1}}}
\newcommand{\papermodelrow}[2]{%
  \multicolumn{#1}{@{}l@{}}{\strut\papermodelname{#2}} \\}
\newcommand{\papermodellabel}[1]{\makebox[0pt][l]{\rule{0pt}{11pt}\small\papermodelname{#1}}}
\newcommand{\papersummaryfirst}[1]{%
  \multicolumn{1}{@{}>{\columncolor{black!6}[0pt][\tabcolsep]}l}{\textbf{#1}}}
\newcommand{\papersummarylast}[1]{%
  \multicolumn{1}{>{\columncolor{black!6}[\tabcolsep][0pt]}c@{}}{#1}}
\newcommand{\papershadedrules}{%
  \setlength{\aboverulesep}{0pt}%
  \setlength{\belowrulesep}{0pt}%
  \renewcommand{\arraystretch}{1.2}}
\newcommand{\oursavg}[1]{\cellcolor{regeOurs!94!black}#1}
\newcommand{\ourslast}[1]{%
  \multicolumn{1}{>{\columncolor{regeOurs!94!black}[\tabcolsep][0pt]}c@{}}{#1}}

\makeatletter
\newcommand{\papermaincaptionstyle}{%
  \long\def\@makecaption##1##2{%
    \vskip\abovecaptionskip
    {\normalfont\normalsize\noindent ##1: ##2\par}%
    \vskip\belowcaptionskip}}
\makeatother

\makeatletter
\newcommand{\paperfixedtablewidth}{%
  \TPT@unhook
  \edef\TPT@hsize{\hsize\the\linewidth\noexpand\@parboxrestore}%
  \TPT@hsize}
\makeatother

\newcommand{\ctl}{Control}
\newcommand{\selfc}{Self-Consistency}
\newcommand{\rege}{ReGE}
\newcommand{\regesel}{\mbox{ReGE-Sel}}
\newcommand{\regesyn}{\mbox{ReGE-Syn}}
\newcommand{\pdr}{PDR}

\newcommand{\regesyna}{\mbox{ReGE-A}}
\newcommand{\agg}{AggAgent}
\newcommand{\reflexion}{Reflexion}

\newcommand{\closedguide}{Closed-book}

\newcommand{\bon}{Best-of-$N$}
\newcommand{\tbb}{Terminal-Bench~2.0}
\newcommand{\spd}{Spider2-Lite}
\newcommand{\aw}{AndroidWorld}
\newcommand{\bcp}{BrowseComp-Plus}
\newcommand{\tableselfc}{SC}

\title{ReGE: Stronger Agents through\\Experience Synthesis}

\author{
{\small \textbf{Jiarong Liang}$^{1}$ \quad
\textbf{Yubo Wang}$^{1}$ \quad
\textbf{Zijie Liu}$^{2}$ \quad
\textbf{Yuxuan Zhang}$^{3}$ \quad
\textbf{Ping Nie}$^{1}$ \quad
\textbf{Wenhu Chen}$^{1}$ }\\[0.5em]
$^{1}$University of Waterloo \quad
$^{2}$University of North Carolina at Chapel Hill \\
$^{3}$University of British Columbia \\[0.4em]
\url{https://github.com/TIGER-AI-Lab/ReGE}
}

\makeatletter
\newcommand{\regearxivmaketitle}{%
  \vbox{\hsize\textwidth\centering
    {\LARGE\upshape\bfseries \@title\par}
    \begin{tabular}[t]{@{}c@{}}\rule{\z@}{24pt}\ignorespaces\@author\end{tabular}\par
    \vskip 0.3in minus 0.1in}}
\makeatother

\newif\ifarxiv
\arxivtrue
\ifarxiv\iclrfinalcopy\hypersetup{hidelinks}\fi
\makeatletter
\ifarxiv\let\@maketitle\regearxivmaketitle\fi
\makeatother
\begin{document}

\maketitle
\ifarxiv\lhead{Preprint}\fi

\begin{abstract}
Repeated attempts at interactive tasks uncover facts, useful operations, and failures that can inform subsequent execution. We introduce ReGE, a framework for turning this recorded experience into guidance for a fresh attempt. Its main implementation, ReGE-Syn, uses a single coach call to jointly synthesize bounded execution records and available artifacts into self-contained guidance, connecting task-specific findings with proposed actions and relevant pitfalls. A student starts from the task's initial environment state and adapts this guidance to current observations. Source generation, synthesis, and student execution use the same base model, without parameter updates or a stronger teacher. Across four benchmarks, ReGE improves accuracy by an average of 9.9 percentage points for GPT and 8.8 percentage points for Claude compared with unguided execution. On matched BrowseComp-Plus subsets, ReGE-Syn reduces guidance-construction tokens by 98.2\% on GPT and 95.5\% on Claude relative to adapted AggAgent, with observed success. Combined construction and execution tokens decrease by 15.7\% and 35.4\%, respectively, counting input plus output and excluding source generation. BrowseComp-Plus ablations further associate access to process information with fewer student steps and show that joint synthesis produces substantially shorter guidance than Direct history or Independent summaries, with observed success. These results demonstrate how agents can use their own recorded experience to improve subsequent task completion, and suggest a low-cost way to consolidate the execution records of parallel agent systems such as agent swarms.
\end{abstract}

\suppressfloats[t]

\section{Introduction}
\label{sec:intro}

\begin{wrapfigure}[17]{r}{0.48\linewidth}
\vspace{-12pt}
\centering
\includegraphics[width=\linewidth]{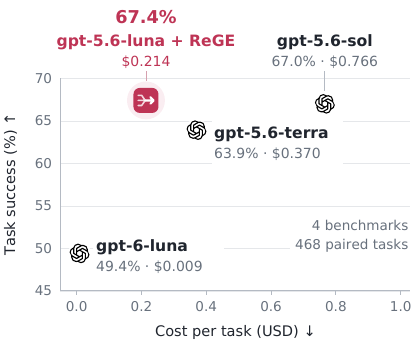}
\setlength{\abovecaptionskip}{0pt}%
\vspace{-12pt}
\caption{\textbf{With \rege{}, \mbox{gpt-5.6-luna} matches \mbox{gpt-5.6-sol} at 28\% of its cost.}
Costs include source pools; see Table~\ref{tab:stronger-models}.}
\label{fig:teaser}
\end{wrapfigure}
Repeated attempts can uncover solutions and information that a single run misses \citep{brown2024monkeys,snell2024scaling}. For interactive agents, these attempts produce more than candidate answers: they record discovered facts, useful tool operations, intermediate results, and failed strategies. This experience offers material for improving the same model's subsequent actions. We ask: \textbf{How can an agent turn experience from its own attempts into more effective subsequent execution?}

Useful discoveries can be distributed across attempts, while the complete records are long and the correctness of individual findings is unknown. Candidate selection, response aggregation, and trajectory summarization provide different ways to reuse prior attempts \citep{cobbe2021verifiers,li2025selfmoa,kim2026stcac}. The challenge is to retain the details needed for subsequent actions while reducing the history a new agent must process. A file created or an application state reached in a previous attempt does not carry over when the environment is reset: the agent must perform the required operations against the current state.

\begin{figure}[t]
\centering
    \includegraphics[width=\textwidth]{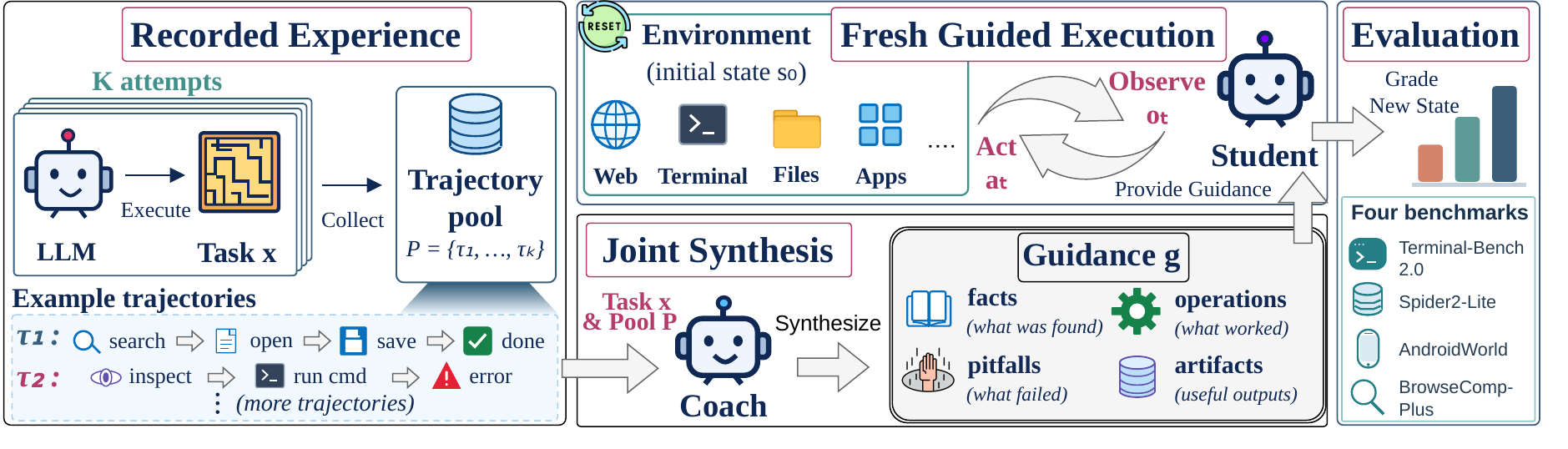}
    
\caption{\textbf{The joint-synthesis implementation of ReGE.} The coach reads the rendered source pool; the student receives the resulting guidance and interacts with its own environment. Source generation, synthesis, and execution share the same base model.}
\label{fig:pipeline}
\end{figure}

We introduce \textbf{ReGE} (\emph{Re-execution Guided by Experience}), a framework for constructing guidance from recorded attempts. Its main implementation, \regesyn{}, presents bounded execution records and available artifacts together to a coach, which produces self-contained guidance in one call. The coach connects task-specific findings with proposed operations and relevant pitfalls. A fresh student starts from the task's initial environment state, using current observations to follow, adjust, and verify the guidance. Source generation, synthesis, and student execution use the same base model, without parameter updates or a stronger teacher; this lets us test whether a lower-priced model guided by its own attempts can match higher-priced models. The framework also includes \regesel{}, which forwards one judge-selected trajectory as guidance (see \S\ref{sec:variants}).

We evaluate ReGE on \tbb{}, \spd{}, \aw{}, and \bcp{} with GPT and Claude model lines. Across four benchmarks, ReGE improves accuracy by an average of 9.9 percentage points for GPT and 8.8 percentage points for Claude compared with unguided execution. On a subset, \mbox{gpt-5.6-luna} with ReGE reaches 67.4\% average success, comparable to the 67.0\% of the higher-priced \mbox{gpt-5.6-sol}, at 28\% of its cost including source-pool generation (Figure~\ref{fig:teaser}). On matched \bcp{} subsets, \regesyn{} reduces guidance-construction tokens by 98.2\% on GPT and 95.5\% on Claude relative to adapted \agg{}, with observed success. Although student execution uses more tokens, combined construction and execution usage falls by 15.7\% and 35.4\%, counting input plus output and excluding source generation. Further \bcp{} ablations examine which source information helps execution, how to organize multiple attempts, and how the guidance's output form affects its use.

Our contributions are:

\begin{itemize}
\item \textbf{Joint synthesis of execution experience.} \regesyn{} connects concrete findings, operations, and failure lessons across recorded attempts into actionable guidance for a fresh execution by the same base model.
\item \textbf{Capability gains with low guidance-construction cost.} Across four benchmarks, ReGE improves mean task completion over unguided execution. On matched \bcp{} subsets, \regesyn{} substantially reduces construction tokens relative to adapted iterative aggregation and retains lower combined usage after student execution. With ReGE, a lower-priced model reaches the success of a higher-priced model in its family at 28\% of its cost.
\end{itemize}

\section{ReGE: Experience-Guided Agent Execution}
\label{sec:framework}

\rege{} is a framework for constructing guidance from recorded experience and using it in a fresh execution. Guidance can be built by jointly synthesizing multiple attempts (\regesyn{}) or selecting one trajectory (\regesel{}). We focus on joint synthesis; \S\ref{sec:variants} compares the variants. The student starts from the task's initial environment state and uses the guidance alongside current observations. Source generation, guidance construction, and student execution share the same base model, without parameter updates or a stronger teacher. Figure~\ref{fig:pipeline} illustrates the synthesis implementation.

\subsection{Recorded Experience}
\label{sec:problem}

A task consists of an instruction $x$, an initial environment state $s_0$, and a benchmark evaluator. Repeated attempts produce a source pool $\mathcal{P}=\{\tau_1,\ldots,\tau_K\}$. For \regesyn{}, we construct the text representation $V(\mathcal{P})$ programmatically: each attempt forms a separate block, with available reasoning or action narratives and commands in step order, followed by final artifact paths and contents. These fields are copied and truncated to bound the input length; no per-trajectory LLM summary is generated before the coach sees them.

Benchmark correctness labels are withheld from both coach and student. Raw per-step observation fields are omitted, so execution feedback is accessible only where preserved in the rendered text or artifacts. This representation provides bounded execution records rather than the complete environment history. Appendices~\ref{app:representation} and~\ref{app:prompts} specify the field mappings, truncation limits, and prompts.

\subsection{Joint Experience Synthesis}
\label{sec:synthesis}

Useful experience can be distributed across attempts: one may uncover a necessary task fact, another a workable command, and another a failure to avoid. \regesyn{} brings these records into a shared context and prompts a coach to turn them into self-contained guidance for a fresh attempt. Using the same base model with parameters $\theta$, the coach $C_\theta$ constructs
\begin{equation}
g=C_\theta\!\left(x,V(\mathcal{P});p_{\mathrm{syn}}\right),
\label{eq:psyn-synthesis}
\end{equation}
where $p_{\mathrm{syn}}$ specifies the guidance construction task.

All retained source blocks are presented together, preserving attempt boundaries and step order. A single coach call produces the guide, allowing decisions about what to retain and connect to account for information across attempts. Independent per-trajectory summarization makes these decisions separately before combining summaries. The coach is prompted to connect concrete findings with proposed operations and relevant pitfalls, retaining details needed to act on them.

The default guidance is an execution-ordered plan. Exact paths, values, commands, format requirements, and encountered problems are placed alongside the steps where they matter. When the records support a candidate answer or artifact, the coach is asked to include it together with a way to verify it. The guide must be understandable without consulting the source attempts, since the student receives only the resulting guidance. Source correctness labels are unavailable, so its recommendations remain subject to the student's checks against the current environment.

\regesyn{} shares joint aggregation with response-aggregation methods such as Self-MoA \citep{li2025selfmoa}. Here, the coach synthesizes bounded execution records and available artifacts into guidance for further interaction: the student must carry out and verify the proposed operations in a fresh environment. Table~\ref{tab:aggregation-designs} compares these design choices; \S\ref{sec:ablation} examines the value of process information, and \S\ref{sec:pdr-ablation} compares joint synthesis with alternative ways of presenting the recorded experience.

\begin{table}[!htbp]
\centering
\papermainstyle
\begin{threeparttable}
\caption{\textbf{Comparison of aggregation methods.}}
\label{tab:aggregation-designs}
\setlength{\tabcolsep}{3pt}
\begin{tabularx}{\linewidth}{@{}l*{6}{>{\centering\arraybackslash}X}@{}}
\toprule
\textbf{Criterion} & \shortstack{SC\\(source)} & Self-MoA & \reflexion{} & \shortstack{Agentic\\PDR} & \agg{} & \textbf{\regesyn{}} \\
\midrule
\textbf{Single-pass Synthesis} & \textcolor{red!75!black}{$\times$} & \textcolor{green!50!black}{$\checkmark$} & \textcolor{red!75!black}{$\times$} & \textcolor{red!75!black}{$\times$} & \textcolor{red!75!black}{$\times$} & \textcolor{green!50!black}{$\boldsymbol{\checkmark}$} \\
\textbf{Trajectory Info.} & \textcolor{red!75!black}{$\times$} & \textcolor{red!75!black}{$\times$} & \textcolor{green!50!black}{$\checkmark$} & \textcolor{green!50!black}{$\checkmark$} & \textcolor{green!50!black}{$\checkmark$} & \textcolor{green!50!black}{$\boldsymbol{\checkmark}$} \\
\textbf{Student Rerun} & \textcolor{red!75!black}{$\times$} & \textcolor{red!75!black}{$\times$} & \textcolor{green!50!black}{$\checkmark$} & \textcolor{green!50!black}{$\checkmark$} & \textcolor{red!75!black}{$\times$} & \textcolor{green!50!black}{$\boldsymbol{\checkmark}$} \\
\textbf{Aggregation Calls} & 0 & 1 & Iterative & $K$ & Iterative & \textbf{1} \\
\bottomrule
\end{tabularx}
\begin{tablenotes}[flushleft]
\papermainnotes
\item Columns describe \selfc{} (\tableselfc{}) source voting \citep{wang2023selfconsistency}, single-pass Self-MoA \citep{li2025selfmoa}, \reflexion{} \citep{shinn2023reflexion}, agentic \pdr{} \citep{kim2026stcac}, and \agg{} \citep{lee2026aggagent}. Aggregation calls count model calls that combine source attempts (reflection calls for \reflexion{}), excluding source generation and task execution; $K$ is the number of source attempts.
\end{tablenotes}
\end{threeparttable}
\end{table}

\subsection{Guided Student Execution}
\label{sec:student}

The student policy $\pi_\theta$ receives the task instruction and guidance, and begins a new interaction from $s_0$:
\begin{equation}
\tau^g\sim\operatorname{Run}(\pi_\theta,x,s_0;g).
\label{eq:guided-execution}
\end{equation}
Files created and environment changes made in the source attempts do not automatically carry over. The student uses its current observations and interaction history to adopt, adjust, or skip the proposed operations, and performs any required actions in this environment. At termination, the benchmark grades the state or submitted output produced by $\tau^g$. Section~\ref{sec:execution-case} illustrates how a student verifies and adapts concrete guidance in a fresh database-recovery execution.

\section{Experimental Setup}
\label{sec:setup}
\label{sec:tasks-models}
\label{sec:arms}
\label{sec:execution-protocol}
\label{sec:evaluation}

\noindent\textbf{Benchmarks.} We evaluate on four benchmarks. \textbf{\tbb{}}~\citep{tbench2025} includes 89 tasks, with a budget of 30 agent turns per execution. \textbf{\spd{}}~\citep{lei2025spider2} includes 135 instances with local SQLite databases; agents receive read-only access and up to 18 SQL calls. \textbf{\aw{}}~\citep{rawles2025androidworld} uses the text-only T3A agent, with guidance supplied through its native task-guideline interface and a step budget of ten times task complexity. After excluding two templates that crash the emulator, we evaluate 114 templates with three parameterisations on GPT and one on Claude, yielding 342 and 114 instances, respectively. \textbf{\bcp{}}~\citep{chen2025browsecompplus} includes 830 questions over a fixed corpus of 100{,}195 documents. Agents use retrieval and document-opening tools without live web access, with a 40-step budget. We use DeepSeek-V4-Flash with reasoning disabled as the LLM equivalence judge for \bcp{}.

\noindent\textbf{Models and budgets.} We use \mbox{gpt-5.6-luna} and \mbox{claude-sonnet-5}, holding the base model fixed across source generation, guidance construction, and student execution. Source pools contain four attempts per task on all benchmarks
(see Appendix~\ref{app:poolcoverage}). \rege{} and \agg{} share these pools and each receive one fresh student execution under the same benchmark-specific step limit to match the budget. Generation settings, effective serving limits, and collection exceptions are documented in Appendix~\ref{app:collection}.

\noindent\textbf{Baselines.} \ctl{} performs one unguided execution (pass@1) under the same student step limit. Our \reflexion{} adaptation~\citep{shinn2023reflexion} allows up to three executions, using reflections from preceding trials and label-free self-evaluation for stopping. \agg{}~\citep{lee2026aggagent} iteratively inspects source records using its released tools and prompts. To align it with our fresh-execution protocol, we adapt its final response into guidance for a fresh student. Independent summaries, our agentic \pdr{} adaptation~\citep{kim2026stcac}, summarize four source trajectories separately and concatenate the summaries to guide a fresh student (Appendix~\ref{app:pdr-protocol}). We also report two references without student re-execution: \selfc{}~\citep[\tableselfc{};][]{wang2023selfconsistency} votes over available source outputs, while \bon{} reports the original score of the source selected by the judge model.

\noindent\textbf{Inference cost.} Our primary cost metric is the sum of input and output tokens used after source-pool generation, including all method-specific guidance construction and execution calls. Input-only usage and costs that include source generation are explicitly identified.

\section{Results}
\label{sec:results}

\begin{table}[ht]
\centering
\papermainstyle
\fontsize{8}{9.5}\selectfont
\setlength{\tabcolsep}{2.3pt}
\papershadedrules
\begin{threeparttable}
\caption{\textbf{Task success and source-inclusive token costs across four benchmarks.}
Cost combines source-generation tokens with measured construction and execution tokens, relative to \ctl{} ($\times$). --- denotes unavailable or inapplicable entries. Shaded columns show our method; bold and underline mark the best and second-best success in each row.}
\label{tab:main}

\newcommand{\methodpart}[1]{{\scriptsize #1}}%
\begin{tabularx}{\linewidth}{@{}X*{5}{c}>{\columncolor{regeOurs}}c*{5}{c}>{\columncolor{regeOurs}[\tabcolsep][0pt]}c@{}}
\toprule
 & \multicolumn{6}{c}{\textbf{Task success (\%)}}
 & \multicolumn{6}{c}{\textbf{Cost ($\times$ \ctl{})}} \\
\cmidrule(lr){2-7}
\cmidrule(l){8-13}
\textbf{Benchmark}
 & \methodpart{\ctl{}} & \methodpart{\tableselfc{}} & \methodpart{\bon{}} & \methodpart{\reflexion{}} & \methodpart{\agg{}} & \methodpart{\rege{}}
 & \methodpart{\ctl{}} & \methodpart{\tableselfc{}} & \methodpart{\bon{}} & \methodpart{\reflexion{}} & \methodpart{\agg{}} & \methodpart{\rege{}} \\
\midrule

\papermodellabel{gpt-5.6-luna} & & & & & & & & & & & & \\
\tbb{}
 & 41.6 & --- & \underline{54.0} & 46.1 & 46.1 & \textbf{56.2}
 & 1.00 & --- & 4.17 & 2.71 & 5.83 & 5.25 \\
Spider2-Lite
 & 59.3 & \underline{64.4} & 63.7 & \underline{64.4} & 55.6 & \textbf{66.7}
 & 1.00 & 4.00 & 4.32 & 1.50 & 11.71 & 5.36 \\
AndroidWorld
 & 53.8 & --- & \underline{57.9} & 55.3 & 51.5 & \textbf{58.8}
 & 1.00 & --- & 4.02 & 2.30 & 4.92 & 5.06 \\
\bcp{}
 & 64.1 & 74.0 & 74.0 & 73.3 & \underline{76.1} & \textbf{76.6}
 & 1.00 & 4.00 & 4.01 & 2.09 & 4.74 & 4.52 \\
\cmidrule{1-13}
\rowcolor{black!6}
\papersummaryfirst{Average}
 & 54.7 & --- & \underline{62.4} & 59.8 & 57.3 & \oursavg{\textbf{64.6}}
 & 1.00 & --- & 4.13 & 2.15 & 6.80 & \ourslast{5.05} \\
\midrule

\papermodellabel{claude-sonnet-5} & & & & & & & & & & & & \\
\tbb{}
 & 40.4 & --- & 38.2 & \underline{48.3} & 46.1 & \textbf{50.6}
 & 1.00 & --- & 4.06 & 2.62 & 6.67 & 5.26 \\
Spider2-Lite
 & 57.8 & 60.0 & \underline{61.5} & 59.3 & 60.7 & \textbf{63.7}
 & 1.00 & 4.00 & 4.16 & 1.48 & 8.75 & 5.04 \\
AndroidWorld
 & 61.4 & --- & 64.9 & 61.4 & \underline{65.8} & \textbf{67.5}
 & 1.00 & --- & 4.01 & 2.07 & 5.09 & 5.02 \\
\bcp{}
 & 74.3 & 84.6 & 84.7 & 84.5 & \underline{86.3} & \textbf{87.5}
 & 1.00 & 4.00 & 4.02 & 2.14 & 4.72 & 4.39 \\
\cmidrule{1-13}
\rowcolor{black!6}
\papersummaryfirst{Average}
 & 58.5 & --- & 62.3 & 63.4 & \underline{64.7} & \oursavg{\textbf{67.3}}
 & 1.00 & --- & 4.06 & 2.08 & 6.31 & \ourslast{4.93} \\
\bottomrule
\end{tabularx}

\end{threeparttable}
\end{table}

\textbf{Recorded experience strengthens execution by the same model.}
Across the four benchmarks, \rege{} improves task completion over
\ctl{} while keeping the base model fixed and requiring a fresh
student execution (Table~\ref{tab:main}). For example, GPT success on \tbb{} rises from
41.6\% to 56.2\%. Here, the student must use prior experience to
perform the required operations in a reset environment, so the
gain reflects improved completion in a new attempt.
Joint synthesis provides a low-cost way to construct such guidance:
on matched settings, \regesyn{} substantially reduces
construction tokens relative to \agg{} while retaining lower
combined construction and student-execution usage, excluding source
generation.

\subsection{Efficient Experience Synthesis}
\label{sec:cost}

\begin{table}[htbp]
\centering
\papermainstyle
\setlength{\tabcolsep}{4pt}
\begin{threeparttable}
\caption{\textbf{Joint synthesis reduces guidance-construction tokens
by over 95\% on both models.}
Results on matched \bcp{} queries.
Token counts are mean input plus output per query, including failures
and excluding source generation.
Construction savings are relative to \agg{}.
See Appendix~\ref{app:accounting} for accounting details.}
\label{tab:cost-matched}

\begin{tabularx}{\linewidth}{
    @{}l*{5}{>{\centering\arraybackslash}X}@{}
}
\toprule
\textbf{Method}
 & \textbf{Success}
 & \multicolumn{2}{c}{\textbf{Construction}}
 & \textbf{Execution}
 & \textbf{Total} \\
\cmidrule(lr){3-4}
 & (\%)
 & Tokens (k)
 & Saved (\%)
 & Tokens (k)
 & Tokens (k) \\
\midrule

\papermodelrow{6}{gpt-5.6-luna}
\agg{}
 & 75.3
 & 232.9
 & ---
 & 44.8
 & 277.7 \\
\regesyn{}
 & 75.2
 & \textbf{4.2}
 & \textbf{98.2}
 & 229.8
 & \textbf{234.1} \\

\midrule
\papermodelrow{6}{claude-sonnet-5}
\agg{}
 & 87.0
 & 211.7
 & ---
 & 134.8
 & 346.6 \\
\regesyn{}
 & 86.2
 & \textbf{9.6}
 & \textbf{95.5}
 & 214.2
 & \textbf{223.8} \\

\bottomrule
\end{tabularx}
\end{threeparttable}
\end{table}

\textbf{Joint synthesis makes effective guidance inexpensive to construct.}
On matched \bcp{} queries, \regesyn{} reduces construction
tokens by more than 95\% relative to \agg{} on both base models,
while maintaining observed task success
(Table~\ref{tab:cost-matched}).
Guidance construction consequently accounts for only a small
fraction of \regesyn{}'s combined usage.
These results show that consolidating prior attempts into useful
guidance need not make construction the dominant inference expense.

\textbf{The construction savings translate into lower combined usage.}
Students guided by \regesyn{} consume more tokens than those guided
by \agg{}, partially offsetting the construction savings.
However, the reduction in construction cost more than compensates
for this additional student work.
On these matched subsets, \regesyn{} uses the fewest combined
construction and execution tokens among \ctl{}, \reflexion{},
\agg{}, and \regesyn{}. For \reflexion{}, construction includes self-evaluation and reflection,
while execution sums token usage across all actor trials.
This comparison counts input plus output tokens and excludes
source generation.

\textbf{Experience guidance can reduce downstream execution effort.}
For \mbox{claude-sonnet-5}, \regesyn{} uses fewer student output tokens
and steps than unguided \ctl{} on \tbb{} and \spd{}
(Table~\ref{tab:student-effort}).
On \spd{}, SQL calls fall by 50\% and output tokens by 30\%,
whereas both measures remain nearly unchanged on \aw{}.
These results extend the efficiency analysis beyond \bcp{}
and show that reductions in student effort vary across tasks.

\begin{table}[htbp]
\centering
\papermainstyle
\begin{threeparttable}
\caption{\textbf{ReGE-Syn reduces student steps and output tokens
on \tbb{} and \spd{}. }
Ratios compare \regesyn{} with unguided \ctl{}. Only student output tokens and steps are counted, excluding input tokens, source generation, judging, and construction.}
\label{tab:student-effort}
\begin{tabularx}{\linewidth}{@{}l*{2}{>{\centering\arraybackslash}X}@{}}
\toprule
\textbf{Benchmark} & \textbf{Student output tokens} & \textbf{Student steps} \\
\midrule
\papermodelrow{3}{claude-sonnet-5\quad\regesyn{} / \ctl{}}
\tbb{} & $0.86\times$ & $0.88\times$ \\
\spd{} & $0.70\times$ & $0.50\times$ \\
\aw{} & $1.02\times$ & $1.01\times$ \\
\bottomrule
\end{tabularx}
\end{threeparttable}
\end{table}

\subsection{Comparison with Other Models}
\label{sec:stronger-models}

\textbf{With \rege{}, a lower-priced model matches higher-priced models of its family at a fraction of their cost.}
We compare \mbox{gpt-5.6-luna} with \rege{} against single unguided executions of two higher-priced models from the same family, \mbox{gpt-5.6-terra} and \mbox{gpt-5.6-sol}, and of the newer \mbox{gpt-6-luna}, which is priced below \mbox{gpt-5.6-luna}.
Due to budget constraints, we use the first 50\% of tasks by task ID on \tbb{}, \spd{}, and \aw{} (task templates with one parameterization each on \aw{}) and the 300 \bcp{} queries used in our ablations.
The other models use the prompts, tools, step limits, and graders of the main \ctl{} runs, and their results are paired by task with the \mbox{gpt-5.6-luna} results of Table~\ref{tab:main}, leaving 468 comparable tasks.
On this subset, \rege{} raises the average task success of \mbox{gpt-5.6-luna} from 56.8\% to 67.4\%, compared with 63.9\% for \mbox{gpt-5.6-terra}, 67.0\% for \mbox{gpt-5.6-sol}, and 49.4\% for \mbox{gpt-6-luna} (Table~\ref{tab:stronger-models} and Figure~\ref{fig:teaser}).
Including source-pool generation, \rege{} costs \$0.21 per task at list prices, 58\% of the cost of \mbox{gpt-5.6-terra} and 28\% of that of \mbox{gpt-5.6-sol}; \mbox{gpt-6-luna} costs \$0.009 per task.
When the source pool already exists, \rege{} costs \$0.034 per task, 9\% and 4\% of the costs of \mbox{gpt-5.6-terra} and \mbox{gpt-5.6-sol} and close to the \$0.041 of one unguided \mbox{gpt-5.6-luna} execution.

\begin{table}[htbp]
\centering
\papermainstyle
\begin{threeparttable}
\caption{\textbf{With \rege{}, \mbox{gpt-5.6-luna} matches \mbox{gpt-5.6-terra} and \mbox{gpt-5.6-sol} at lower cost.}
Task success (\%) and list-price cost per task (USD) on the task subset of \S\ref{sec:stronger-models}.
The other models run once per task without guidance; the cost of \rege{} is shown including source-pool generation and with an existing source pool.
Shaded columns show our method. Bold marks the best value in each row, and underline marks the second best.}
\label{tab:stronger-models}
\setlength{\tabcolsep}{4.5pt}%
\papershadedrules
\newcommand{\modelpart}[1]{{\scriptsize\renewcommand{\arraystretch}{1}#1}}%
\begin{tabularx}{\linewidth}{@{}Xc>{\columncolor{regeOurs}}cccc>{\columncolor{regeOurs}}c>{\columncolor{regeOurs}}cccc@{}}
\toprule
 & \multicolumn{5}{c}{\textbf{Task success (\%)}}
 & \multicolumn{5}{c}{\textbf{Cost per task (USD)}} \\
\cmidrule(lr){2-6}
\cmidrule(l){7-11}
 & \multicolumn{4}{c}{\modelpart{gpt-5.6-}}
 & \modelpart{gpt-6-}
 & \multicolumn{2}{c}{\modelpart{gpt-5.6-luna\,+\,\rege{}}}
 & \multicolumn{2}{c}{\modelpart{gpt-5.6-}}
 & \modelpart{gpt-6-} \\
\cmidrule(lr){2-5}
\cmidrule(lr){6-6}
\cmidrule(lr){7-8}
\cmidrule(lr){9-10}
\cmidrule(l){11-11}
\textbf{Benchmark (tasks)}\rule[-3.3ex]{0pt}{0pt}
 & \modelpart{luna}
 & \modelpart{\begin{tabular}[t]{@{}c@{}}luna\\+\,\rege{}\end{tabular}}
 & \modelpart{terra} & \modelpart{sol} & \modelpart{luna}
 & \modelpart{\begin{tabular}[t]{@{}c@{}}incl.\\pool\end{tabular}}
 & \modelpart{\begin{tabular}[t]{@{}c@{}}existing\\pool\end{tabular}}
 & \modelpart{terra} & \modelpart{sol} & \modelpart{luna} \\
\midrule
\tbb{} (44)
 & 40.9 & \underline{63.6} & 59.1 & \textbf{68.2} & 34.1
 & 0.187 & \underline{0.028} & 0.251 & 0.601 & \textbf{0.005} \\
\spd{} (68)
 & \underline{66.2} & \textbf{77.9} & \underline{66.2} & 63.2 & \underline{66.2}
 & 0.043 & \underline{0.007} & 0.044 & 0.090 & \textbf{0.001} \\
\aw{} (56)
 & 50.0 & 48.2 & \underline{55.4} & \textbf{57.1} & 44.6
 & 0.313 & \underline{0.062} & 0.568 & 1.138 & \textbf{0.025} \\
\bcp{} (300)
 & 70.0 & \textbf{80.0} & 75.0 & \underline{79.3} & 52.7
 & 0.314 & \underline{0.038} & 0.615 & 1.237 & \textbf{0.004} \\
\cmidrule{1-11}
\rowcolor{black!6}
\papersummaryfirst{Average}
 & 56.8 & \oursavg{\textbf{67.4}} & 63.9 & \underline{67.0} & 49.4
 & \oursavg{0.214} & \oursavg{\underline{0.034}} & 0.370 & 0.766 & \papersummarylast{\textbf{0.009}} \\
\bottomrule
\end{tabularx}
\end{threeparttable}
\end{table}

\textbf{The newer, lower-priced \mbox{gpt-6-luna} costs less than \rege{} but is less accurate.}
\rege{} exceeds \mbox{gpt-6-luna} by 18.1 points on average. \rege{} is higher on all four benchmarks, though by only 3.6 points on \aw{}.
\mbox{gpt-6-luna} costs 4\% as much as \rege{} including source-pool generation and 26\% as much with an existing source pool.
Appendix~\ref{app:stronger-models} details task selection and prices.

\subsection{What Information Makes Experience Useful?}
\label{sec:realization}
\label{sec:cond}
\label{sec:ablation}
\label{sec:source-diagnostics}

\textbf{Recorded experience makes coaching more effective.}
With \mbox{claude-sonnet-5} on the same 300 \bcp{} queries, coaching without
source experience performs to \ctl{}, while providing
even one recorded attempt substantially improves success
in Table~\ref{tab:source-count}.
Increasing the source count from one to four adds another
4.0 percentage points.
Source coverage and input length increase together, so this
comparison does not establish gains at matched inference cost.
Appendix~\ref{app:ksweep} provides the paired comparisons.

\begin{table}[htbp]
\centering
\papermainstyle
\begin{threeparttable}
\caption{\textbf{Task success increases as the coach receives
more source attempts.}
\ctl{} has no coach; \closedguide{} uses the coach with zero
source attempts.
Coached configurations share the same coach, synthesis request,
delivery header, and requested output budget;
$K'=1,2,4$ expose prefixes of the same ordered source pools.}
\label{tab:source-count}

\begin{tabularx}{\linewidth}{
    @{}l*{5}{>{\centering\arraybackslash}X}@{}
}
\toprule
\textbf{Configuration}
 & \shortstack{\ctl{}\\No coach}
 & \shortstack{\closedguide{}\\$K'=0$}
 & \shortstack{\regesyn{}\\$K'=1$}
 & \shortstack{\regesyn{}\\$K'=2$}
 & \shortstack{\regesyn{}\\$K'=4$} \\
\midrule
Task success (\%)
 & 74.7 & 75.7 & 84.3 & 86.3 & \textbf{88.3} \\
\bottomrule
\end{tabularx}
\end{threeparttable}
\end{table}

\textbf{Process records help students complete tasks with less
downstream work.}
Table~\ref{tab:fao} compares final-artifact-only (FAO) synthesis with
full-input \regesyn{} on the same queries.
Queries with only one submitted source answer (6.3\% of all queries)
account for 90\% of the net additional successes.
The execution traces also reveal a completion bottleneck:
on the queries solved only with full input (5.7\% of all queries), 94.1\% of FAO runs
exhaust the 40-step budget without submitting an answer.
Even when all four sources submit answers, success rates
remain nearly identical while mean student steps fall from
8.9 to 4.8.
Thus, the additional information supports both task completion
and more efficient execution, including when final answers
are already available.

\begin{table}[htbp]
\centering
\papermainstyle
\begin{threeparttable}
\caption{\textbf{Adding process records yields higher task success
with fewer student steps.}
FAO uses final artifacts only; \regesyn{} additionally receives
process records.
Both conditions share the source pools, coach, plan request,
and student protocol.}
\label{tab:fao}

\begin{tabularx}{\linewidth}{
    @{}l*{2}{>{\centering\arraybackslash}X}@{}
}
\toprule
\textbf{Metric / source-answer group}
 & \textbf{FAO}
 & \textbf{\regesyn{}} \\
\midrule

\multicolumn{3}{@{}l}{\textit{Overall performance}} \\
Task success (\%) $\uparrow$
 & 85.0
 & \textbf{88.3} \\
Mean student steps $\downarrow$
 & 14.6
 & \textbf{9.1} \\

\midrule
\multicolumn{3}{@{}l}{
    \textit{Task success (\%), grouped by submitted source answers}
} \\
0 source answers (10.0\% of queries) & 10.0 & 6.7 \\
1 source answer (6.3\% of queries) & 47.4 & 94.7 \\
2 source answers (8.3\% of queries) & 84.0 & 96.0 \\
3 source answers (8.7\% of queries) & 96.2 & 96.2 \\
4 source answers (66.7\% of queries) & 98.5 & 98.0 \\

\bottomrule
\end{tabularx}
\end{threeparttable}
\end{table}

\subsection{When Does Recorded Experience Help?}
\label{sec:pool-quality}

\textbf{ReGE-Syn shows its largest gains on pools containing
both successful and failed attempts.}
Table~\ref{tab:pool-quality} compares \regesyn{} with
\ctl{} within source-quality groups.
Mixed pools show the largest observed success-rate gains
in all eight model--benchmark combinations.
Across \tbb{}, \spd{}, and \aw{}, these gains range
from 13.0 to 21.6 percentage points.
On \bcp{}, they reach 48.0 points on GPT and 35.2 on Claude.
These results show that source experience need not be uniformly
successful to support more effective subsequent execution.

\begin{table}[t]
\centering
\papermainstyle
\begin{threeparttable}
\caption{\textbf{ReGE-Syn achieves its largest observed success-rate gains
 over \ctl{} on mixed source pools.}
Tasks are grouped by whether none, some, or all source attempts succeed.
Entries report fresh student success rates in percent;
Tasks (\%) is the percentage of the benchmark's tasks that fall in each group.
Source correctness labels are used only for analysis and are
unavailable to the coach and student.}
\label{tab:pool-quality}
\label{tab:strata}
\newcommand{\methodpart}[1]{{\scriptsize #1}}%
\begin{tabularx}{\linewidth}{@{}ll*{6}{>{\centering\arraybackslash}X}@{}}
\toprule
 & & \multicolumn{3}{c}{\papermodelname{gpt-5.6-luna}} & \multicolumn{3}{c}{\papermodelname{claude-sonnet-5}} \\
\cmidrule(lr){3-5}\cmidrule(l){6-8}
\textbf{Benchmark} & \shortstack[l]{\textbf{Source-pool}\\\textbf{condition}} & \methodpart{Tasks (\%)} & \methodpart{\ctl{}} & \methodpart{\regesyn{}} & \methodpart{Tasks (\%)} & \methodpart{\ctl{}} & \methodpart{\regesyn{}} \\
\midrule
\tbb{} & All failed & 36.0 & 0.0 & 9.4 & 52.8 & 12.8 & 14.9 \\
 & Mixed & 41.6 & 51.4 & 73.0 & 25.8 & 52.2 & 65.2 \\
 & All succeeded & 22.5 & 90.0 & 100.0 & 21.3 & 94.7 & 100.0 \\
\addlinespace[0.65ex]
\spd{} & All failed & 24.4 & 0.0 & 0.0 & 26.7 & 2.8 & 0.0 \\
 & Mixed & 32.6 & 52.3 & 72.7 & 25.9 & 48.6 & 62.9 \\
 & All succeeded & 43.0 & 98.3 & 100.0 & 47.4 & 93.8 & 100.0 \\
\addlinespace[0.65ex]
\aw{} & All failed & 36.0 & 9.8 & 10.6 & 30.7 & 8.6 & 8.6 \\
 & Mixed & 26.9 & 57.6 & 75.0 & 25.4 & 62.1 & 79.3 \\
 & All succeeded & 37.1 & 93.7 & 91.3 & 43.9 & 98.0 & 98.0 \\
\addlinespace[0.65ex]
\bcp{} & All failed & 23.9 & 5.1 & 7.6 & 14.0 & 6.0 & 12.1 \\
 & Mixed & 20.6 & 43.9 & 91.8 & 23.6 & 59.2 & 94.4 \\
 & All succeeded & 55.5 & 97.0 & 99.6 & 62.4 & 95.6 & 99.8 \\
\bottomrule
\end{tabularx}
\end{threeparttable}
\end{table}

\textbf{Source quality also reveals the limits of experience reuse.}
Fresh successes remain possible when all source attempts fail,
but improvements over \ctl{} are smaller and inconsistent.
On GPT \tbb{}, success rises from 0.0\% to 9.4\%,
whereas both methods remain at 0.0\% on GPT \spd{}.
When all sources succeed, \ctl{} already solves at least 90\%
of tasks in every model--benchmark combination,
leaving less room for improvement.
Appendix~\ref{app:source-quality} examines re-execution conditioned
on a successful selected source, and Appendix~\ref{app:repeat}
reports repeated-execution results.

\subsection{Joint Synthesis for Compact Guidance}
\label{sec:pdr-ablation}

\textbf{Joint synthesis produces shorter guidance with
observed success.}
With \mbox{claude-sonnet-5} on the 300 matched \bcp{} queries in
Table~\ref{tab:pdr-cost}, \regesyn{} produces guidance roughly
one quarter as long as Direct history or Independent summaries.
Combined construction and student-execution tokens decrease
by 19.8\% and 31.8\%, respectively, excluding source generation.

\begin{table}[htbp]
\centering
\papermainstyle
\begin{threeparttable}
\caption{\textbf{Joint synthesis reduces guidance length and
total token usage while achieving observed success.}
Total tokens sum construction and student input plus output,
including cache reads/writes and construction retries,
and exclude source generation.}
\label{tab:pdr-cost}

\begin{tabularx}{\linewidth}{
    @{}l*{3}{>{\centering\arraybackslash}X}@{}
}
\toprule
\textbf{Method}
 & \shortstack{\textbf{Success}\\(\%) $\uparrow$}
 & \shortstack{\textbf{Guidance length}\\(characters) $\downarrow$}
 & \shortstack{\textbf{Total tokens}\\$\downarrow$} \\
\midrule
Direct history
 & \textbf{88.3}
 & 20,571
 & 268.9k \\
Independent summaries
 & 86.9
 & 19,590
 & 316.5k \\
\regesyn{}
 & \textbf{88.3}
 & \textbf{5,063}
 & \textbf{215.8k} \\
\bottomrule
\end{tabularx}
\end{threeparttable}
\end{table}

All methods reuse the same four-source pools for a fresh student.
Direct history concatenates rendered trajectories without a coach;
Independent summaries concatenate
independently generated source summaries;
\regesyn{} constructs one guide jointly.
Independent summaries share source content, delivery header,
and student protocol with \regesyn{}.
Direct history uses a different rendering and header, so its
comparison evaluates complete reuse strategies.
Requested output limits are 2,500 tokens per independent summary
and 10,000 for joint synthesis; these do not imply equal
realised usage.
Appendix~\ref{app:pdr-protocol} gives prompts and retries.

\subsection{Guidance Variants}
\label{sec:variants}

\textbf{Joint synthesis supports effective guidance beyond a fixed
execution plan.}
\regesyna{} requests an explanation and answer while retaining
\regesyn{}'s coach, source pool, and delivery header.
On the same 290 queries in
Table~\ref{tab:variants}, this change reduces combined construction
and student-execution tokens by 32.1\%, while observed success
rises from 88.3\% to 90.7\%.
Most of the savings come from student execution, with additional
construction savings mainly from shorter coach output.
The answer format does not lead in every tested setting:
in a separate comparison with \mbox{gpt-5.6-luna} on 100 \bcp{}
queries with the same source pools and protocol, \regesyna{} solves
80\% of queries versus 82\% for \regesyn{}.

\textbf{Selecting one trajectory provides effective guidance.}
As shown in Table~\ref{tab:variants}, \regesel{} achieves 89.3\% success
compared with 88.3\% for \regesyn{}, while using 21\% fewer student tokens.

\begin{table}[htbp]
\centering
\papermainstyle
\begin{threeparttable}
\caption{\textbf{Explanation-and-answer guidance lowers token usage
with higher observed task success.}
Total is construction plus student execution, excluding source pool
generation.}
\label{tab:variants}
\label{tab:guidance-strategies}

\begin{tabularx}{\linewidth}{
    @{}l
    >{\raggedright\arraybackslash}p{0.235\linewidth}
    *{4}{>{\centering\arraybackslash}X}@{}
}
\toprule
\textbf{Variant}
 & \textbf{Guidance form}
 & \shortstack{\textbf{Success}\\(\%) $\uparrow$}
 & \shortstack{\textbf{Guidance}\\\textbf{construction} $\downarrow$}
 & \shortstack{\textbf{Student}\\\textbf{execution} $\downarrow$}
 & \textbf{Total} $\downarrow$ \\
\midrule

\papermodelrow{6}{
    claude-sonnet-5\quad\bcp{}
}

\regesel{}
 & Selected trajectory
 & 89.3
 & $0.86\times$
 & $0.79\times$
 & $0.80\times$ \\

\regesyna{}
 & Explanation + answer
 & \textbf{90.7}
 & $\mathbf{0.85}\times$
 & $\mathbf{0.67}\times$
 & $\mathbf{0.68}\times$ \\

\regesyn{}
 & Execution plan
 & 88.3
 & $1.00\times$
 & $1.00\times$
 & $1.00\times$ \\

\bottomrule
\end{tabularx}
\end{threeparttable}
\end{table}

\Needspace{12\baselineskip}
\subsection{From Recorded Experience to Fresh Execution}
\label{sec:execution-case}

\textbf{Concrete guidance supports verification and adaptation during execution.}
A \mbox{claude-sonnet-5} run on the \tbb{} task \texttt{db-wal-recovery} illustrates how
a prior finding and a failure lesson guide operations against
the student's current environment.

\begin{swenextbox}{Execution Case: Recovering a Database from Its Log}
\small
\textbf{Recorded experience.}
Successful source attempts identify the XOR decoding key
\texttt{0x42} for the database's write-ahead log.
Another attempt opens \texttt{main.db} before making a backup
and loses \texttt{main.db-wal}, leaving the recovery incomplete.

\medskip
\textbf{Synthesized guidance.}
The guide retains the decoding key and instructs the student
to back up both files before opening the database, decode and
check the log, and then recover the records. The case illustrates guidance that preserves concrete information
and a student that checks and adjusts the proposed operations.

\medskip
\textbf{Student verification and adjustment.}
The student backs up its current files, verifies the key and
decoded log, and moves the \texttt{main.db-shm} cleanup ahead
of replacing the log and opening the database.
The guide had suggested this cleanup only as a fallback. The student exports 11 ordered records to \texttt{recovered.json}
and passes all seven official tests.

\end{swenextbox}

\Needspace{10\baselineskip}
\textbf{Incomplete attempts can still supply useful operational knowledge.}
In the \aw{} task \texttt{BrowserMultiply}, none of the four
\mbox{claude-sonnet-5} source attempts completes the task, but two reach a working route
for opening the local web page. Here the sources supply an access procedure rather than a completed
answer; the student must still obtain the page values and finish
the interaction. 

\begin{swenextbox}{Execution Case: Completing a Browser Task from Unsuccessful Attempts}
\small
\textbf{Recorded experience.}
Repeated taps on \texttt{task.html} leave the file manager unchanged.
Two attempts discover the route long-press $\rightarrow$ More options
$\rightarrow$ Open with $\rightarrow$ Chrome, but all four exhaust
their 22-step budgets before completing the multiplication task.

\medskip
\textbf{Synthesized guidance.}
The guide retains this opening sequence, warns against repeated
single taps, and describes Chrome's initial setup prompts.
It asks the student to obtain the five numbers from the page
and submit their product.

\medskip
\textbf{Student verification and adjustment.}
The student uses the opening sequence immediately.
When dismissing Chrome's sync prompt returns it to the file manager,
it repeats Open with $\rightarrow$ Chrome to reach the page.
It records the displayed numbers $9,10,9,5,5$ during the five clicks
and enters their product, $20{,}250$. The page displays ``Success!'' and the benchmark records success
after 19 steps, within the same 22-step budget as the sources.

\end{swenextbox}

\section{Related work}
\label{sec:related}

\noindent\textbf{Aggregation of multiple attempts.} Repeated sampling produces candidates that can be selected through voting \citep{wang2023selfconsistency} or evaluation \citep{cobbe2021verifiers}, or combined through generative aggregation \citep{jiang2023llmblender,wang2024moa}. Self-MoA jointly aggregates responses sampled from a single model \citep{li2025selfmoa}. \agg{} extends aggregation to long-horizon agentic tasks by using tools to inspect candidate solutions and retrieve information from their trajectories \citep{lee2026aggagent}. \regesyn{} shares the goal of combining useful information across attempts. It jointly reads bounded execution records and available artifacts in one coach call, producing guidance that a student uses during a fresh interaction with the environment. Agent swarms decompose a task into heterogeneous sub-problems that parallel sub-agents execute concurrently \citep{kimi2026k25}; \regesyn{} studies the complementary step of consolidating the records of parallel attempts at the same task into guidance for one fresh execution.

\noindent\textbf{Distillation for subsequent execution.} Parallel-Distill-Refine constructs a workspace for subsequent generations using operators that include joint global summarization and candidate selection \citep{madaan2025improvement}. \citet{kim2026stcac} adapt PDR to agentic coding by independently summarizing rollouts and concatenating selected summaries to condition new rollouts in freshly initialized environments. \regesyn{} constructs guidance jointly from the retained source records, allowing content choices to account for information across attempts before producing the guide.

\noindent\textbf{Reflection and agent memory.} \reflexion{} uses verbal reflections on task feedback to guide subsequent trials \citep{shinn2023reflexion}. ExpeL extracts insights from accumulated experience \citep{zhao2024expel}, while Agent Workflow Memory induces reusable workflows \citep{wang2024awm}. AutoGuide derives context-aware guidelines from offline experience \citep{fu2024autoguide}, and AutoManual organizes knowledge acquired through interaction into instruction manuals \citep{chen2024automanual}. Voyager stores executable skills for later reuse \citep{wang2024voyager}; A-MEM organizes experience into linked notes and revises existing memories as new information arrives \citep{xu2025amem}. We study task-specific reuse, asking how records from multiple attempts at the same problem can support a new execution with the same base model.

\section{Limitations and Conclusion}
\label{sec:conclusion}

Due to computational budget constraints, our evaluation is limited to four benchmarks: Terminal-Bench 2.0, Spider2-Lite, AndroidWorld, and BrowseComp-Plus, with ablation studies conducted on selected subsets of the evaluation data. Further evaluation should cover broader task settings and repeated executions with aligned settings.

We presented ReGE, a framework that turns recorded attempts into guidance for fresh agent execution. Its main implementation jointly synthesizes concrete findings, operations, and pitfalls in a single coach call. Across four benchmarks, experience-guided execution improves mean task completion while keeping the base model fixed. On matched \bcp{} subsets, joint synthesis sharply reduces guidance-construction tokens relative to adapted iterative aggregation and lowers combined construction and execution usage, excluding source generation. \bcp{} analyses further support compact experience reuse through multiple guidance forms. These findings establish a practical way to make an agent's prior attempts useful for subsequent action.

\subsection*{AI Use Statement}
We used generative AI tools to assist with language polishing and textual revisions of the manuscript, including improvements to clarity, grammar, and presentation. The authors take responsibility for the final content of the paper.

\subsection*{Ethics statement}
The benchmarks used are public research benchmarks with no human subjects; \aw{} runs in an emulator and \bcp{} uses a fixed offline corpus, so no real user accounts are accessed. The methods reuse an agent's own attempts and do not require additional data collection. We see no direct potential for harm beyond that of stronger autonomous agents in general.

\bibliography{iclr2027_conference,psyn}
\bibliographystyle{iclr2027_conference}

\clearpage
\appendix
\setlength{\textfloatsep}{10pt plus 2pt minus 2pt}
\setlength{\floatsep}{8pt plus 2pt minus 2pt}
\setlength{\intextsep}{6pt plus 2pt minus 2pt}
\FloatBarrier
\section{Experimental protocols}
\label{app:protocols}

\subsection{Source Representation}
\label{app:representation}

\textbf{Fields and artifacts.} Artifacts are produced files on \tbb{}, submitted SQL and query results on \spd{}, submitted answers on \bcp{}, and action narratives and the final action summary on \aw{}. Empty or non-submitting attempts remain in the pool.

\textbf{Truncation and overflow.} \regesyn{}, \regesel{} and \regesyna{} retain at most 120 rendered steps per source, with reasoning and commands truncated to 200 and 160 characters per step, respectively. Each artifact is capped at 6,000 characters. If the combined source serialization exceeds 350,000 characters, all artifact bodies are removed while their paths remain. Any remaining excess is truncated at that limit and followed by an explicit marker.

Direct history uses a step-only rendering that omits tool observations and separate final-artifact blocks.

\subsection{Prompts and Guidance Delivery}
\label{app:prompts}

The boxes reproduce construction prompts and delivery text, with placeholders and line breaks normalized for typesetting.

\begin{swenextbox}{ReGE-Syn: Joint Synthesis Prompt}
\small
\textbf{System message.}
You are an aggregation agent: several independent agents attempted a task (correctness unknown) and you synthesize the single best guidance for a fresh agent. You do NOT have access to the ground truth solution.

\textbf{User message.}

TASK: $\langle$instruction$\rangle$

$K$ independent previous attempts (full trajectories + final files):

$\langle$serialized source pool$\rangle$

Write an EXECUTABLE STEP-BY-STEP PLAN for a fresh agent that will solve the task from scratch: (1) numbered steps in execution order, each concrete enough to act on; (2) the critical facts the attempts uncovered -- exact paths, values, commands, formats, edge cases -- placed at the steps where they matter; (3) pitfalls the attempts exposed and how to avoid them. If the evidence already supports a final answer or artifact, state it in the plan together with how to verify it. The plan must be self-contained: the fresh agent sees only this plan, never the attempts.
\end{swenextbox}

\begin{swenextbox}{ReGE-A: Alternative Construction Request}
\small
Produce your solution as a single string with exactly two XML sections: \texttt{<explanation>...</explanation>}\allowbreak\texttt{<answer>...</answer>}. The explanation must be self-contained -- a reader who never saw this prompt understands how the answer was derived. The answer is your best final solution to the task.
\end{swenextbox}

\begin{swenextbox}{ReGE-Sel: Trajectory Selection Prompt}
\small
\textbf{System message.} You judge agent trajectories.

\textbf{User message.}

TASK: $\langle$instruction$\rangle$

$K$ independent agents attempted this task (success labels hidden):

$\langle$rendered source trajectories$\rangle$

Which ONE rollout is most likely FULLY correct end-to-end? Reply with EXACTLY one rollout id (e.g.\ claude-sonnet\_1) and nothing else.
\end{swenextbox}

Up to eight judge attempts seek a parsable rollout ID, with requested output limits given in Table~\ref{tab:generation-settings}. Parsing prefers an exact ID, then the last valid ID mentioned; if all attempts fail, it selects the first source without consulting correctness labels. Cached picks avoid repeating selection during student collection.

\begin{swenextbox}{Student Delivery Headers}
\small
\textbf{Coach outputs (\regesyn{}, \regesyna{}, and \closedguide{}).}

\#\# GUIDANCE (synthesized by an aggregation agent that inspected several previous independent attempts at this task; follow it but verify as you go):

\textbf{Selected trajectory.}

\#\# A PREVIOUS AGENT'S COMPLETE ATTEMPT (judged most likely correct among several independent attempts; follow it but verify as you go):

(chosen: $\langle$rollout ID$\rangle$, label-free judge)

\end{swenextbox}

Guidance precedes the instruction on \tbb{} and \spd{}, enters the native T3A task-guideline interface on \aw{}, and is appended to the user message on \bcp{}. The experiments use same-model source pools.

\begin{swenextbox}{AndroidWorld: Reset and Verification Instructions}
\small
\textbf{GPT header and instruction.}

\#\# RESET-AND-VERIFICATION GUIDANCE

The following material comes from a previous attempt at this task. Your environment has been reset to its initial state. Use current observations to determine which actions are needed, and verify the current task state before declaring completion.

\textbf{Claude instruction, following the selected-trajectory header.}

The following record comes from a separate, earlier execution. The environment for your current execution has been reset to the task's initial state. Statements about actions taken or completion in the record describe that earlier execution only; they do not establish the state or completion of your current task.
\end{swenextbox}

\textbf{Independent-summary construction.}
\label{app:pdr-protocol}
Each independent-summary call receives the task and one block from \regesyn{}'s bounded pool serialization, inheriting its truncation and artifact-redaction decisions. The reconstructed prompt requests one JSON object with four groups: task requirements, agent actions, artifact changes, and command execution. Initial length instructions allow approximately 1,000 words, six command entries, six items per list, and 250 characters per string, prioritizing concrete commands, paths, values, and failure causes.

Up to three formatting attempts retain the first complete JSON object containing the required task-requirement and agent-action keys. Retries request 500 words, three commands, four items per list, and 120 characters per string; the output-token request remains fixed at the value in Table~\ref{tab:generation-settings}. If none passes, the last nonempty response is kept and flagged; an all-empty response becomes a placeholder. There is no quality-based selection or fallback to raw trajectories.
Usage counts all recorded calls for each delivered summary, including its original generation when a cached summary is reused.

\subsection{Evaluation and Token Accounting}
\label{app:accounting}

\textbf{Serving and collection conditions.}
\label{app:collection}
Table~\ref{tab:generation-settings} distinguishes requested limits from serving evidence. The comparisons measure realized usage under method-specific construction and stopping procedures.

\begin{table}[!htbp]
\centering
\papermainstyle
\begin{threeparttable}
\caption{\textbf{Requested generation limits and available serving evidence.}}
\label{tab:generation-settings}
\setlength{\tabcolsep}{4pt}
\renewcommand{\arraystretch}{1.12}
\begin{tabularx}{\linewidth}{@{}>{\raggedright\arraybackslash}p{0.28\linewidth}>{\raggedright\arraybackslash}p{0.23\linewidth}>{\raggedright\arraybackslash}X@{}}
\toprule
\textbf{Component} & \textbf{Output-token request} & \textbf{Scope / evidence} \\
\midrule
\regesyn{}, \regesyna{}, \closedguide{} & 10,000 per call & Shared construction implementation. \\
\regesel{} judge & 2,000 / 8,000 / 20,000 & Attempts 1--2 / 3--4 / 5--8, respectively. \\
\agg{} & 10,000 per round & Default limit of 100 aggregation rounds. \\
Independent summaries & 2,500 per source & Matched PDR adaptation; Appendix~\ref{app:pdr-protocol}. \\
Claude \tbb{} student & 12,000 per call & 30 agent turns \\
GPT \tbb{} student & 12,000 per call & 30 agent turns \\
\spd{}/\aw{} students & No explicit cap & Main collector request paths omit an output limit. \\
\bcp{} student & 4,000 per call & Main collector request. \\
\bottomrule
\end{tabularx}
\end{threeparttable}
\end{table}

\FloatBarrier
\section{Source pools and repeated execution}
\label{app:strata}

\subsection{Source-pool Characteristics}
\label{app:poolcoverage}
\label{app:ksweep}

Table~\ref{tab:source-diagnostics} reports source-pool coverage.

\begin{table}[H]
\centering
\papermainstyle
\begin{threeparttable}
\caption{\textbf{Source-pool coverage.}}
\label{tab:source-diagnostics}
\begin{tabularx}{\linewidth}{@{}l*{2}{>{\centering\arraybackslash}X}@{}}
\toprule
\textbf{Benchmark} & \textbf{GPT coverage (\%)} & \textbf{Claude coverage (\%)} \\
\midrule
\tbb{} & 64.0 & 47.2 \\
\spd{} & 75.6 & 73.3 \\
\aw{} & 64.0 & 69.3 \\
\bcp{} & 76.1 & 85.9 \\
\bottomrule
\end{tabularx}
\begin{tablenotes}[flushleft]
\papermainnotes
\item Coverage is the fraction of tasks with at least one successful source attempt, using all source pools. Pool sizes are specified in \S\ref{sec:setup}; later student analyses use common valid execution records.
\end{tablenotes}
\end{threeparttable}
\end{table}

\textbf{Six-source experiment.} With \mbox{claude-sonnet-5} on a separate 100-query \bcp{} subset, student success is 81\%, 82\%, 85\%, and 89\% with one, two, four, and six sources, respectively, versus 70\% for \ctl{} and 87\% source-pool coverage. Six versus four gives $+4/{-}0$ ($p=.125$).

\subsection{Re-execution from a Successful Selected Source}
\label{app:source-quality}

\textbf{Success of the selected source.} Table~\ref{tab:cond} reports fresh student success conditional on the judge selecting a successful source; the same judge pick defines the subset for every method, including those that do not use it.

\begin{table}[H]
\centering
\papermainstyle
\begin{threeparttable}
\caption{\textbf{Fresh student success given a successful selected source.}
Tasks (\%) is the percentage of the benchmark's tasks on which the judge selects a successful source;
the other columns report fresh student success rates in percent.}
\label{tab:cond}
\begin{tabularx}{\linewidth}{@{}lc*{4}{>{\centering\arraybackslash}X}@{}}
\toprule
\textbf{Benchmark} & \textbf{Tasks (\%)} & \ctl{} & \agg{} & \regesel{} & \regesyn{} \\
\midrule
\papermodelrow{6}{gpt-5.6-luna}
\tbb{} & 52.8 & 68.1 & 78.7 & \textbf{97.9} & 89.4 \\
\spd{} & 63.7 & 89.5 & 79.1 & 95.3 & \textbf{96.5} \\
\aw{} & 58.8 & 82.1 & 76.6 & 72.6 & \textbf{87.6} \\
\bcp{} & 74.0 & 83.9 & 98.4 & \textbf{100.0} & 98.7 \\
\midrule
\papermodelrow{6}{claude-sonnet-5}
\tbb{} & 38.2 & 79.4 & 91.2 & \textbf{100.0} & 85.3 \\
\spd{} & 61.5 & 86.7 & 89.2 & \textbf{95.2} & \textbf{95.2} \\
\aw{} & 64.9 & 87.8 & 93.2 & \textbf{94.6} & 91.9 \\
\bcp{} & 84.8 & 85.6 & 99.1 & \textbf{99.7} & 99.0 \\
\bottomrule
\end{tabularx}
\end{threeparttable}
\end{table}

\subsection{Repeated Student Executions}
\label{app:repeat}

On the 35 \spd{} tasks with mixed four-source pools, \ctl{} and \regesyn{} each have three \mbox{claude-sonnet-5} student executions (Table~\ref{tab:repeat}) with fixed source experience and guidance settings.

\begin{table}[H]
\centering
\papermainstyle
\begin{threeparttable}
\caption{\textbf{Three \mbox{claude-sonnet-5} student executions on 35 mixed-pool \spd{} tasks.}}
\label{tab:repeat}
\begin{tabularx}{\linewidth}{@{}l*{2}{>{\centering\arraybackslash}X}@{}}
\toprule
\textbf{Metric} & \ctl{} & \regesyn{} \\
\midrule
Mean success (\%) & 48.6 & 63.8 \\
Tasks solved on all three executions (\%) & 25.7 & 60.0 \\
Tasks solved on at least one execution (\%) & 71.4 & 65.7 \\
\bottomrule
\end{tabularx}
\begin{tablenotes}[flushleft]
\papermainnotes
\item Mean success averages all three executions; the other two rows are percentages of the same 35 tasks.
\end{tablenotes}
\end{threeparttable}
\end{table}

\FloatBarrier
\section{Student Execution Accounting}
\label{app:cost}
\label{app:student-effort}

Table~\ref{tab:student-effort} sums recorded model output tokens
over each student execution. Steps count trace entries on
\tbb{} (including the initial snapshot), SQL calls on \spd{},
and recorded agent steps on \aw{}.
Ratios divide method means on the common valid task set,
rather than averaging per-task ratios.
The calculation uses output tokens only.

\FloatBarrier
\section{Comparison with Other Models}
\label{app:stronger-models}

\textbf{Task selection.} Section~\ref{sec:stronger-models} uses the first 44 of 89 \tbb{} tasks in alphabetical order of task name, the first 68 of 135 \spd{} instances in order of instance ID, and the first 56 of 114 \aw{} templates in alphabetical order of template name. Each \aw{} template uses its first parameterization, the one also used for \mbox{claude-sonnet-5}. On \bcp{}, the comparison uses the 300 queries of our ablations. The selection depends only on task order, not on any results. 

\textbf{Execution.} \mbox{gpt-5.6-terra}, \mbox{gpt-5.6-sol}, and \mbox{gpt-6-luna} each run once per task without guidance, with the prompts, tools, step limits, and graders of the main \mbox{gpt-5.6-luna} \ctl{} runs. Each model uses the serving default-medium effort. 

\textbf{Cost accounting.} List prices in USD per million input and output tokens are 0.20 and 1.20 for \mbox{gpt-5.6-luna}, 2.00 and 12.00 for \mbox{gpt-5.6-terra}, 4.00 and 20.00 for \mbox{gpt-5.6-sol}, 0.10 and 0.50 for \mbox{gpt-6-luna}.

\FloatBarrier
\section{Additional Execution Cases and Evidence}
\label{app:execution-case}

The cases below use recorded \mbox{claude-sonnet-5} executions and the same
source--guidance--student structure as \S\ref{sec:execution-case}.

\subsection{Evidence for the Main-text Cases}
\label{app:case-checks}

Table~\ref{tab:execution-case} documents the starting conditions
and record linkage for the two main-text cases.

\begin{table}[H]
\centering
\papermainstyle
\begin{threeparttable}
\caption{\textbf{Record checks for the main-text execution cases.}}
\label{tab:execution-case}
\begin{tabularx}{\linewidth}{@{}>{\raggedright\arraybackslash}p{0.22\linewidth}>{\raggedright\arraybackslash}X@{}}
\toprule
\textbf{Check} & \textbf{Recorded evidence} \\
\midrule
Database initial state & Initial database and log fingerprints match across the source attempts and student; source-created backups and recovered outputs are not inherited. \\
\addlinespace[0.5ex]
Browser task instance & Source and student task parameters, both random seeds, and the 22-step budget match. The collector resets and initializes each task. Saved summaries record the opening sequence, product entry, and success message. \\
\addlinespace[0.5ex]
Record linkage & Student records lack an injected-guidance hash; linkage uses saved guides, construction/execution logs, and collector configurations. \\
\bottomrule
\end{tabularx}
\end{threeparttable}
\end{table}

\Needspace{12\baselineskip}
\subsection{Adapting a Database Query to Stored Values}
\label{app:case-spider}

\spd{} task \texttt{local098} asks how many actors have
no four-year interval without a film credit between their first
and last credited years.

\begin{swenextbox}{Execution Case: Checking Year Formats before Counting Actors}
\small
\textbf{Recorded experience.}
The four sources return different counts after varying how they
parse \texttt{Movie.year} and bound the gaps between credited years.
Some extract the last four characters; others cast or filter the
raw year strings.

\medskip
\textbf{Synthesized guidance.}
The guide retains distinct actor--year pairs and a \texttt{LAG}
calculation, with consecutive credited years at most four years
apart. It also recommends filtering for raw four-digit year strings
and asks the student to inspect the stored format first.

\medskip
\textbf{Student verification and adjustment.}
The student queries nonstandard year strings and their four-character
suffixes, then checks for nulls and invalid suffixes.
Its final query uses \texttt{CAST(substr(year,-4) AS INTEGER)}
instead of the guide's strict raw-string filter, while retaining
the proposed gap calculation.

\medskip
\textbf{Execution outcome.}
The submitted query returns $28{,}699$ actors, matching an accepted
official result. The record contains ten SQL calls.
\end{swenextbox}

The student retains the query structure while adapting preprocessing
to the data. Intermediate calls retain SQL prefixes and status flags;
the final query and result CSV are complete.

\Needspace{12\baselineskip}
\subsection{Reusing Document Leads for Fresh Verification}
\label{app:case-bcp}

\bcp{} query \texttt{bcp-285} asks for the student who
succeeded a scientist as director of a research center, using
clues about the center, the scientist, and a co-edited publication.

\begin{swenextbox}{Execution Case: Following a Research Lead through Source Documents}
\small
\textbf{Recorded experience.}
One attempt reaches the Citrus Experiment Station and Herbert
J.\ Webber, then submits Leon Dexter Batchelor.
Two other attempts reach relevant documents without submitting
an answer; the remaining attempt spends its budget on other leads.

\medskip
\textbf{Synthesized guidance.}
The guide retains the institution and person names, document IDs
\texttt{9606} and \texttt{26507}, and a reminder to read beyond
the first excerpt. It prioritizes this lead and lists clues that
still require checking.

\medskip
\textbf{Student verification and adjustment.}
The student searches for the institution, reads document
\texttt{9606} at offsets 0 and 6,000, and opens Webber's biography
in document \texttt{26507}. These passages provide the succession
and biographical evidence; an additional search checks the
University of Nebraska's charter date.

\medskip
\textbf{Execution outcome.}
The student submits ``Leon Dexter Batchelor,'' which receives
an exact-match success grade, after six recorded tool steps
including submission.
\end{swenextbox}

The guide preserves both a candidate answer and the document
locations that let the student check the research lead.

\end{document}

%% file: math_commands.tex
\usepackage{amsmath,amsfonts,bm}

\def\eqref#1{equation~\ref{#1}}

\def\1{\bm{1}}

\DeclareMathAlphabet{\mathsfit}{\encodingdefault}{\sfdefault}{m}{sl}
\SetMathAlphabet{\mathsfit}{bold}{\encodingdefault}{\sfdefault}{bx}{n}

